\documentclass[journal]{IEEEtran}

\usepackage{cite}
\usepackage{amsmath,amssymb,amsfonts}
\usepackage{graphicx}
\usepackage{caption}
\usepackage{enumitem}
\usepackage{subcaption}
\usepackage{textcomp}
\usepackage{xcolor}
\usepackage{booktabs}
\usepackage{siunitx}
\usepackage[hidelinks]{hyperref}

\begin{document}

\title{A Bi-level Co-design Framework for Lunar Monoped with 2D Jump and Landing Control}

\author{%
Sanjay Ramkumar$^{1}$\thanks{$^{1}$ Dept. of Aerospace Engineering, Indian Institute of Space Science and Technology. Email: sanjayramkumar27@gmail.com}
\and
Suraj Kumar$^{2}$\thanks{$^{2}$ Controls and Digital Area, 
U R Rao Satellite Center. Email: surajk@ursc.gov.in}
}

\maketitle

\begin{abstract}
Legged robots for lunar exploration must be designed for low gravity where reduced contact forces and long flight phases make the leg morphology and controller choice interdependent. We study this for a planar monoped in lunar gravity (\SI{1.625}{m/s^2}) with a bi-level co-design framework: a differential evolution outer loop searches over link lengths, and each candidate is evaluated on a vertical jump by either a Quadratic program based task-space controller (QP-TSC) or a direct-collocation based trajectory optimization. The two controllers differ on the optimal link lengths : trajectory optimization outperforms the QP-TSC across most of the design space (by 17\%  in Cost of Transport (COT) at their respective optimums). We then extend to a 2D jump, where trajectory optimization with centroidal angular momentum regulation produces stable in-place and forward jumps, and an LQR controller absorbs impact and stabilizes the robot after touchdown, demonstrated on a baseline morphology (both links 0.3 m).
\end{abstract}


\section{Introduction}
\label{sec:intro}
Legged robots are a candidate for exploring lunar terrain where wheeled rovers cannot reach, and in low gravity the most efficient gaits are those with long flight phases, as in hopping and pronking~\cite{8794136,8967552}. This same property makes control harder. When flight phases are long relative to stance, attitude cannot be regulated through ground reaction forces alone, so angular momentum acquired at take-off must be handled during flight to avoid a crash landing~\cite{Rudin_2022}.

The central question we try to answer is whether the optimal leg morphology of a lunar hopper depends on the controller used to evaluate it. We study it in a vertical (1D) jump of a planar monoped in lunar gravity. A bi-level framework searches over link lengths with differential evolution, and each candidate is evaluated by one of two inner controllers: a QP-TSC or a direct-collocation trajectory optimizer.

A vertical jump hides the problems that dominate in practice: pitch, horizontal motion, and landing. We therefore also develop a planar (2D) jump and landing controller. A trajectory optimization that regulates centroidal angular momentum produces in-place and forward jumps, and an LQR controller absorbs the impact and stabilizes the robot after touchdown in simulation. We characterize the landing's region of attraction empirically over touchdown conditions, since formal estimation for hybrid legged systems is difficult because of contact resets~\cite{choi2022computationregionsattractionhybrid}.

\section{Related Work}
\label{sec:related}

\subsection{Low Gravity Legged and Hopping Robots}
Hopping has long been proposed for planetary mobility because ballistic motion is energy efficient in reduced gravity. Recent legged platforms build on this. SpaceBok uses a reaction wheel and
parallel springs for lunar gravity jumping~\cite{8967552}. SpaceHopper and Olympus use learned policies for in-flight reorientation~\cite{spiridonov2024spacehoppersmallscaleleggedrobot,olsen2025olympusjumpingquadrupedplanetary}. Closest to our planar setting, Hari et~al.\ simulate a bipedal hopper with a torso reaction wheel and a PID flight controller in lunar gravity~\cite{hari2026dynamicmodelingattitudecontrol}.

\subsection{Co-design of Legged Robots}

Co-design optimizes the mechanical parameters like link lengths, spring stiffness or actuator selection together with control rather than fixing one and optimizing the other. The two are coupled : a morphology performs well with a controller that exploits it and a controller can only achieve what its body allows.

Prior work includes morphology optimized jointly with motions or policies through sensitivity analysis~\cite{Ha-RSS-17}, joint policy learning~\cite{schaff2018jointlylearningconstructcontrol}, and differentiation through a motion planner~\cite{dinev2022versatilecodesignapproachdynamic}. Bravo-Palacios et~al.\ co-optimize a nominal trajectory, a feedback policy and the morphology under disturbances, and report better performance than designs whose feedback controller is designed separately~\cite{10.1115/1.4052463}. Several works target monoped jumping directly. Fadini et~al.\ use a genetic algorithm outer loop with a trajectory-optimization inner problem and report large energy savings~\cite{9560988}. Singh et~al.\ propose a three stage framework that jointly maximizes jump height and minimizes mechanical energy, with detailed actuator and gearbox modeling~\cite{11203144}. For low gravity jumpers, link lengths and spring stiffness have been chosen by grid search under Mars gravity~\cite{olsen2023martianlavatubeexploration}. 

\subsection{Landing Stability and Regions of Attraction}
Computing regions of attraction for legged systems is difficult because contact causes discontinuous state resets~\cite{choi2022computationregionsattractionhybrid}. Sum of squares methods give formal estimates but can be conservative for highly nonlinear systems~\cite{osti_10182618}, and have been used for balance and step recovery~\cite{Posa-RSS-17}. We use an LQR landing controller and characterize its region of attraction empirically, by sweeping touchdown conditions in the nonlinear simulation.

This study fixes the task and outer loop and vary only the inner controller, to isolate how controller choice affects the optimal lunar monoped morphology and demonstrate
a stable 2D jump and landing.

The remainder of the paper is organized as follows - Section \ref{sec:model} defines the robot model. Section \ref{sec:framework} explains the bi-level co-design framework. Section \ref{sec:res1d} presents the results obtained from the 1D jump co-design. Section \ref{sec:planar} explains the controller design for 2D jump case and finally Section \ref{sec:res2d} presents the results for the 2D jump. Section \ref{sec:disc} and \ref{sec:concl} summarizes the major results from the paper and concludes the paper.

\section{System Model}
\label{sec:model}

The Monoped consists of 3 links - base, thigh and shank.
The state variables for the 1D case are $\mathbf{q} = [z,\ q_1,\ q_2]^T$, while for the 2D case they are $\mathbf{q} = [z,\ x,\ \theta,\ q_1,\ q_2]^T$.
Here, \(z\) denotes the vertical height of the robot base, \(x\) denotes the horizontal displacement of the robot base, \(\theta\) denotes the pitch of the robot base, \(q_1\) denotes the Joint~1 (hip) angle, and \(q_2\) denotes the Joint~2 (knee) angle.

\subsection{Robot Model}
\label{sec:model:robot}

\begin{figure}[h!]
    \centering
    \includegraphics[width=0.5\linewidth]{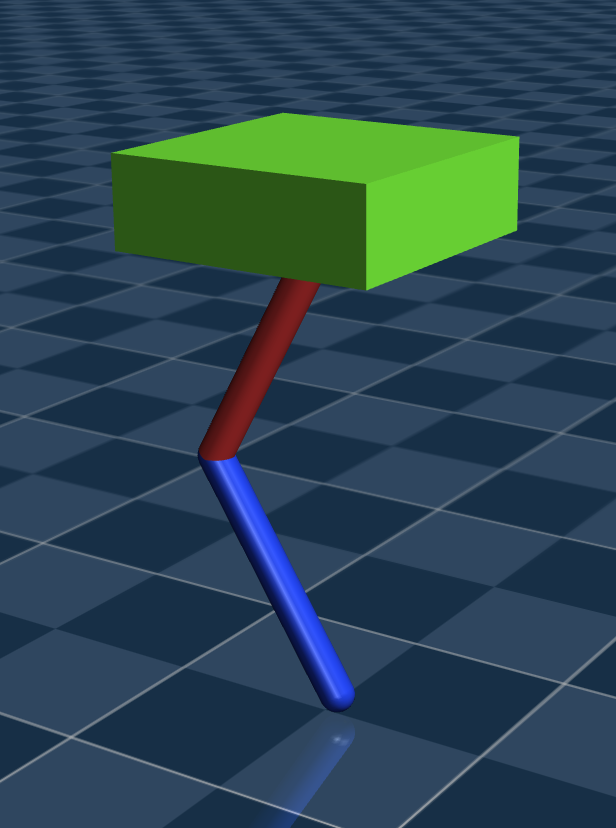}
    \caption{Robot built in MuJoCo}
    \label{fig:model}
\end{figure}

The leg is a planar 2-link leg with the thigh and shank chosen as the co-design variables having a constant density of \SI{1989.4}{kg/m^3} (radius = 0.02 m).  This allows for automatic mass and inertia scaling during the co-design. The base of the robot is modeled as a cuboid of dimensions - 0.15 x 0.15 x 0.05 m and fixed mass of 3 kg. The actuators are direct-drive ideal actuators with torque limits set to 25 Nm. 

\subsection{Robot Dynamics}
\label{sec:model:dyn}
The planar monoped is described by the generalized coordinates
\begin{equation}
\mathbf q =
\begin{bmatrix}
z & q_1 & q_2
\end{bmatrix}^{T},
\end{equation}
with dynamics
\begin{equation}
\mathbf M(\mathbf q)\ddot{\mathbf q}
+\mathbf h(\mathbf q,\dot{\mathbf q})
=
\mathbf S^T\boldsymbol{\tau}
+\mathbf J_c^T\boldsymbol{\lambda},
\end{equation}
where the actuation selection matrix is
\begin{equation}
\mathbf S=
\begin{bmatrix}
0&1&0\\
0&0&1
\end{bmatrix}.
\end{equation}

\section{Bi-level Co-design Framework}
\label{sec:framework}

\begin{figure}[h!]
    \centering
    \includegraphics[width=1\linewidth]{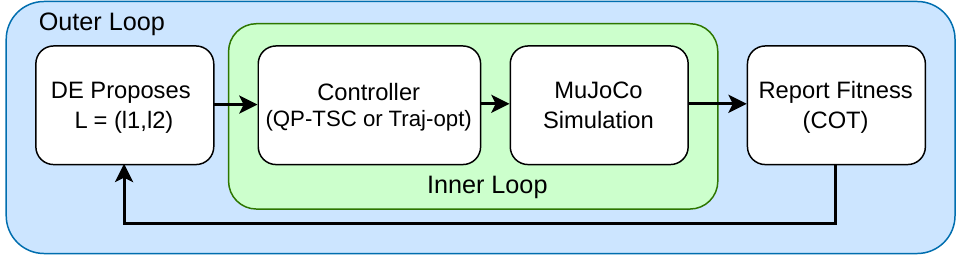}
    \caption{Bi-Level framework}
    \label{fig:framework}
\end{figure}

We pose the co-design problem as a bi-level optimization. The outer loop searches over the thigh and shank lengths $(l_1, l_2)$. For each candidate, an inner controller attempts the \SI{0.2}{m} jump, and the resulting cost of transport is returned to the outer loop as the fitness. The initial hip angle is fixed at $45^\circ$ for all morphologies, and the initial knee angle is solved so that the center of mass lies above the foot (static equilibrium).

\subsection{Differential Evolution Outer Loop}
\label{sec:framework:outer}

The outer loop uses differential evolution (DE), a population based optimizer that needs only function values. Gradients of the COT with respect to link lengths are hard to obtain, because each evaluation runs a MuJoCo simulation of a single jump which could fail. A gradient free search was therefore needed.

DE keeps a population of candidate designs $(l_1, l_2)$. In each generation, a mutant is built for every member by adding a scaled difference of other members to a randomly chosen base member, with scale factor $F$. Exponential crossover with rate $C_r$ mixes the mutant with its parent, and the resulting trial design replaces the parent if its COT is lower. Designs that fail the task, fail the controller, or violate static stability receive a penalty, so they are not selected. The full settings are in the Appendix~\ref{app:imp:de} .






\subsection{Performance Metric}
\label{sec:framework:metric}
The task of the monoped is to achieve a jump height equal to 0.2 m in lunar gravity.
The objective of this Co-Design is to identify the morphologies that achieve the given task with highest energy efficiency. But using raw actuator energy alone would bias the optimizer towards smaller link lengths (Fig.\ref{fig:enerqp} and Fig.\ref{fig:enertraj}) as mass and inertia scale down with them. Thus to make the comparison between different morphologies fair, the Cost of Transport is used as the cost function. COT is defined as - 
\[
\mathrm{COT}
=
\frac{E}{mgh},
\]

where \(E\) is the total positive mechanical work performed by the actuators during the stance phase, \(m\) is the total robot mass, \(g\) is the gravitational acceleration, and \(h\) is the jump height, defined as the change in the center of mass (COM) height from liftoff to the apex.

\subsection{Inner Controller 1: QP based Task Space Controller}
\label{sec:framework:qp}
To achieve the required jump height the liftoff velocity can be calculated using
\[
v_{\mathrm{liftoff}}
=
\sqrt{2gh_{des}}.
\]
The Task space controller tracks a given reference trajectory by minimizing a quadratic cost on deviation from the reference trajectory subject to constraints. During stance, the goal is to achieve a vertical COM velocity equal to $v_{liftoff}$. During flight phase the task is to prepare for landing and hold landing configuration. During the landing phase, the task is to absorb the impact and come to a nominal pose.

A phase-dependent reference trajectory is then generated for the
stance, flight, and landing phases. During stance, the desired COM
acceleration is obtained using
\begin{equation}
\ddot z_{\mathrm{des}}
=
K_p(z_{\mathrm{ref}}-z)
+K_d(\dot z_{\mathrm{ref}}-\dot z)
+\ddot z_{\mathrm{ref}}.
\end{equation}

At each control timestep, the QP optimizes
\begin{equation}
\mathbf x=
\begin{bmatrix}
\ddot{\mathbf q}\\
\boldsymbol{\tau}\\
\boldsymbol{\lambda}
\end{bmatrix}
\end{equation}
using the objective
\begin{equation}
\min_{\mathbf x}\;
\frac{1}{2}
\left\|
\mathbf A\mathbf x-\mathbf \ \ddot{\mathbf x}_{\mathrm{des}}.
\right\|_{\mathbf W}^{2}
+\frac{1}{2}\mathbf x^T\mathbf R\mathbf x,
\end{equation}
where
\begin{equation}
\ddot x_{\mathrm{des}}
=
K_p(x_{\mathrm{ref}}-x)
+K_d(\dot x_{\mathrm{ref}}-\dot x)
+\ddot x_{\mathrm{ref}}.
\end{equation}
which is written in standard QP form as
\begin{equation}
\min_{\mathbf x}\;
\frac{1}{2}\mathbf x^T\mathbf H\mathbf x
+\mathbf f^T\mathbf x,
\end{equation}
where
\begin{equation}
\mathbf H=\mathbf A^T\mathbf W\mathbf A+\mathbf R,
\qquad
\mathbf f=-\mathbf A^T\mathbf W\mathbf \ \ddot{\mathbf x}_{\mathrm{des}}.
\end{equation}

The COM acceleration constraint is
\begin{equation}
\mathbf J_{\mathrm{COM}}\ddot{\mathbf q}
=
\ddot z_{\mathrm{des}}
-\dot{\mathbf J}_{\mathrm{COM}}\dot{\mathbf q}.
\end{equation}

During stance, the QP additionally enforces the fixed-foot constraint
\begin{equation}
\mathbf J_c\ddot{\mathbf q}
+\dot{\mathbf J}_c\dot{\mathbf q}
=\mathbf 0,
\end{equation}
together with the friction and actuator constraints
\begin{equation}
\lambda_z\geq0,
\qquad
|\lambda_x|\leq\mu\lambda_z,
\qquad
|\tau_i|\leq\tau_{\max}.
\end{equation}

Thus, the stance-phase QP is compactly expressed as
\begin{equation}
\begin{aligned}
\min_{\mathbf x}\quad&
\frac12\mathbf x^T\mathbf H\mathbf x+\mathbf f^T\mathbf x\\
\mathrm{s.t.}\quad&
\mathbf M\ddot{\mathbf q}+\mathbf h
=\mathbf S^T\boldsymbol{\tau}
+\mathbf J_c^T\boldsymbol{\lambda},\\
&
\mathbf J_c\ddot{\mathbf q}
+\dot{\mathbf J}_c\dot{\mathbf q}=0,\\
&
\lambda_z\geq0,\qquad
|\lambda_x|\leq\mu\lambda_z,\\
&
|\boldsymbol{\tau}|\leq\tau_{\max}.
\end{aligned}
\end{equation}

During flight, the contact force is removed,
$\boldsymbol{\lambda}=\mathbf0$, giving
\begin{equation}
\mathbf M\ddot{\mathbf q}+\mathbf h
=\mathbf S^T\boldsymbol{\tau},
\qquad
|\boldsymbol{\tau}|\leq\tau_{\max}.
\end{equation}

\subsection{Inner Controller 2: Trajectory Optimization}
\label{sec:framework:trajopt}

The jumping motion is formulated as a constrained nonlinear trajectory optimization problem
over the stance phase. CasADi along with IPOPT was used to solve the optimal control problem. Flight phase and landing phase are handled by joint space PD control. 

The optimization minimizes the positive mechanical work performed by the actuators.

The instantaneous actuator power is

\[
P
=
\boldsymbol{\tau}^T
\dot{\mathbf{q}}_a,
\]

where

\[
\dot{\mathbf{q}}_a=
\begin{bmatrix}
\dot q_1\\
\dot q_2
\end{bmatrix}.
\]

Since only positive work contributes to energy expenditure, a smooth approximation of

\[
\max(P,0)
\]

using the softplus function is used for better convergence,

\[
P^{+}
=
max(P,0)+
\epsilon
\ln
\left(
1+
e^{-|P|/\epsilon}
\right),
\]

where

\[
\epsilon=0.1.
\]

The total stance energy is

\[
E
=
\sum_{k=1}^{N}
P_k^{+}
\Delta t.
\]

Thus the problem is formulated as - 

\begin{equation}
\begin{aligned}
\min_{\mathcal{X}}\quad & E \\[2pt]
\text{s.t.}\quad
& \mathbf{M}(\mathbf q)\ddot{\mathbf q}
 + \mathbf{h}(\mathbf q,\dot{\mathbf q})
 = \mathbf{S}^{T}\boldsymbol{\tau}
 + \mathbf{J}_c^{T}\boldsymbol{\lambda},
 \\[2pt]
& V_{\mathrm{liftoff}}=\sqrt{2gh_{\min}},
 \\[2pt]
& \mathbf q_0=\mathbf q_{\mathrm{nom}},
 \qquad
  \dot{\mathbf q}_0=\mathbf 0,
 \\[2pt]
& \mathbf q_{k+1}
 =\mathbf q_k+
 \frac{\Delta t}{2}
 \left(\dot{\mathbf q}_k+\dot{\mathbf q}_{k+1}\right),
 \\[2pt]
& \dot{\mathbf q}_{k+1}
 =\dot{\mathbf q}_k+
 \frac{\Delta t}{2}
 \left(\ddot{\mathbf q}_k+\ddot{\mathbf q}_{k+1}\right),
 \\[2pt]
& \mathbf J_c\ddot{\mathbf q}
 +\dot{\mathbf J}_c\dot{\mathbf q}
 =\mathbf 0,
 \\[2pt]
& \lambda_z\geq0,
 \qquad
 -\mu\lambda_z\leq\lambda_x\leq\mu\lambda_z,
 \\[2pt]
& -\tau_{\max}\leq\boldsymbol{\tau}
 \leq\tau_{\max}.
\end{aligned}
\label{eq:trajectory_optimization}
\end{equation}

The constraints are -  robot dynamics, the jump height constraint, the initial configuration, trapezoidal collocation, the fixed foot constraint, the friction cone and positive normal force constraints, and the actuator torque limits.

\section{Results: 1D Co-design}

\label{sec:res1d}

\subsection{Landscapes and Optima}
\label{sec:res1d:landscape}

\begin{figure*}[ht]
    \centering
    \begin{subfigure}[t]{0.32\textwidth}
        \centering
        \includegraphics[width=\linewidth]{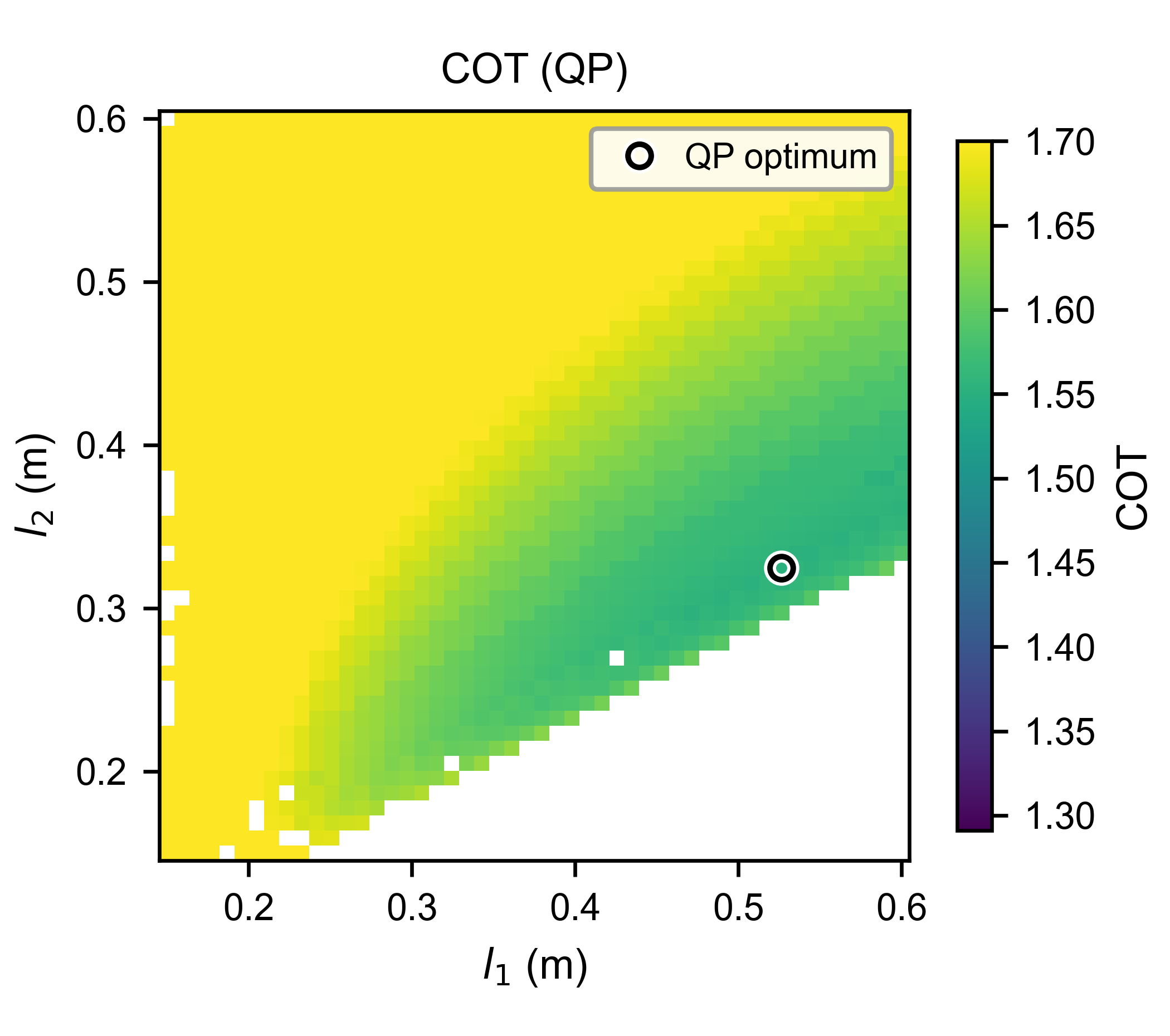}
        \caption{QP-TSC}
        \label{fig:cot_qp}
    \end{subfigure}\hfill
    \begin{subfigure}[t]{0.32\textwidth}
        \centering
        \includegraphics[width=\linewidth]{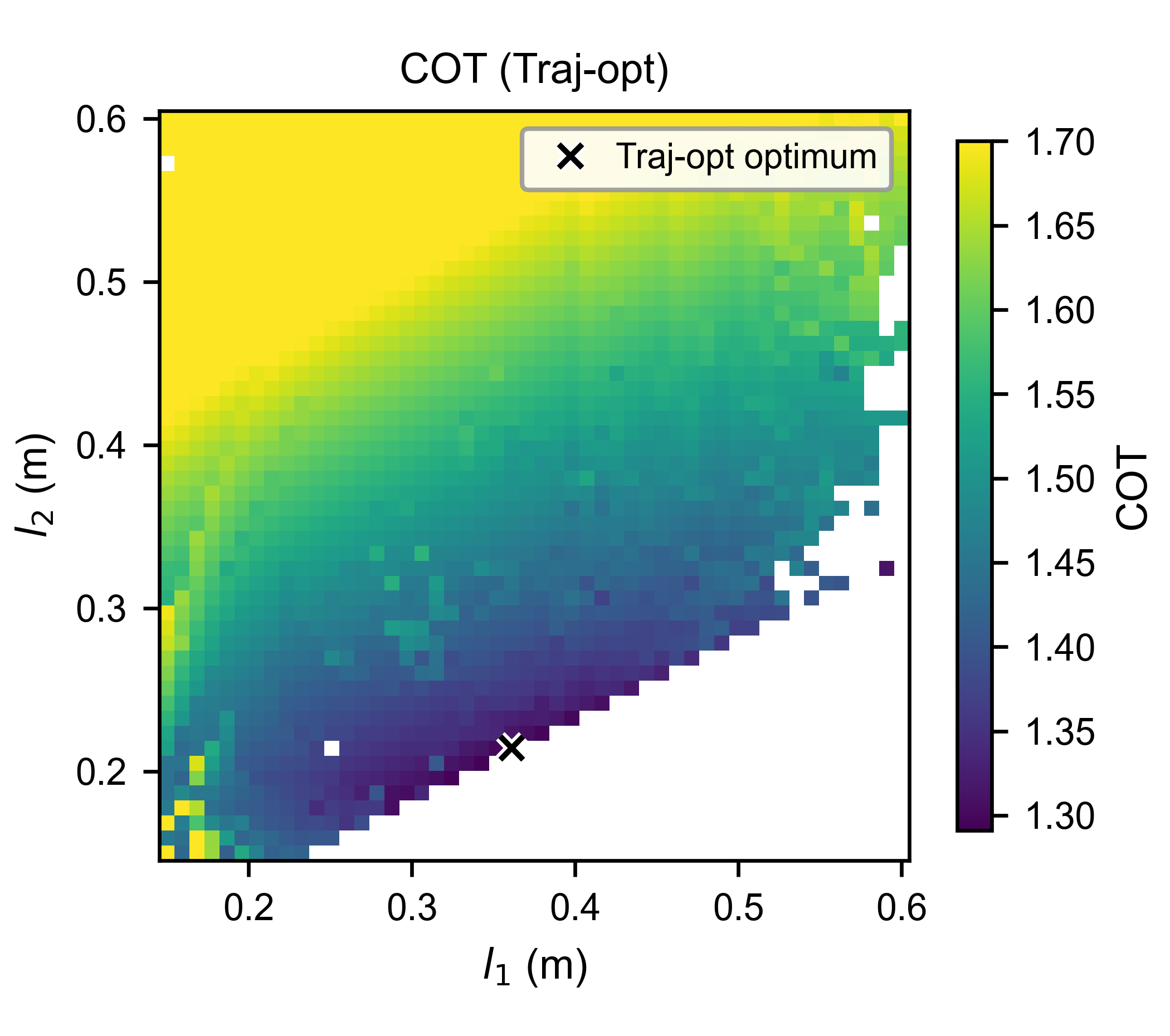}
        \caption{Trajectory optimization}
        \label{fig:cot_traj}
    \end{subfigure}\hfill
    \begin{subfigure}[t]{0.32\textwidth}
        \centering
        \includegraphics[width=\linewidth]{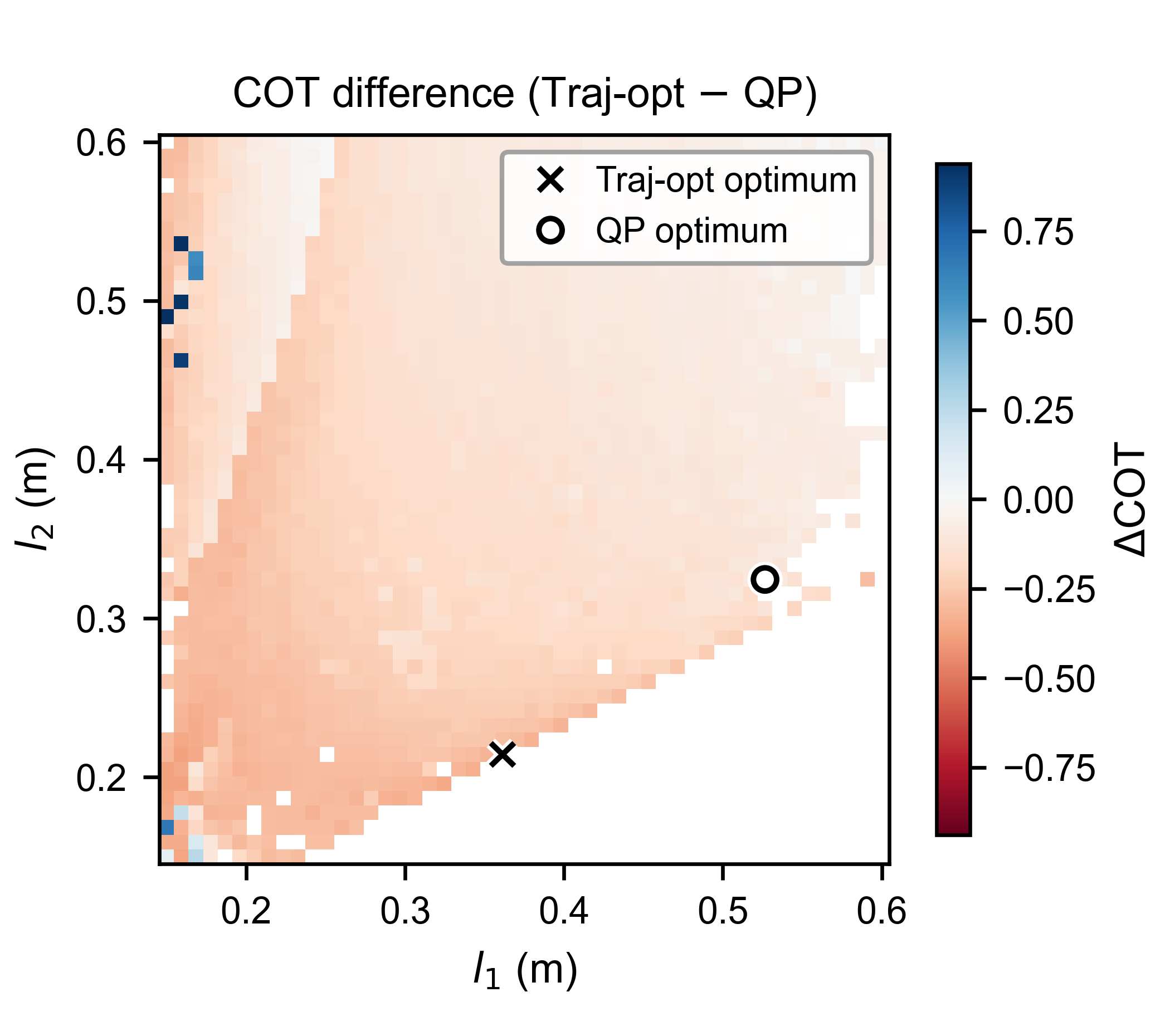}
        \caption{COT difference}
        \label{fig:cot_diff}
    \end{subfigure}
    \caption{COT landscapes over link lengths for the two controllers, and the difference between them.}
    \label{fig:cot_all}
\end{figure*}

The optimum is reported as the mean of 5 independent DE runs with different seed and compared with the baseline morphology - 

\begin{table}[h]
\centering
\caption{Optimal designs from five independent DE runs}
\label{tab:de_runs}
\resizebox{\columnwidth}{!}{%
\begin{tabular}{|l|c|c|c|c|c|}
\hline
\textbf{Controller} & $\mathbf{l_1}$ (m) & $\mathbf{l_2}$ (m)
& \textbf{Opt. COT} & \textbf{Base COT} & \textbf{Imp. (\%)} \\
\hline
QP-TSC & 0.54 & 0.34 & 1.55 & 1.65 & 6.06 \\
\hline
Traj-opt & 0.36 & 0.21 & 1.28 & 1.57 & 18.47 \\
\hline
\end{tabular}%
}
\end{table}

\subsection{Cross-Evaluation}
\label{sec:res1d:cross}

\begin{table}[h]
\centering
\caption{Comparison of COT for Different Design-Controller pairs.}
\label{tab:design_controller_comparison}
\begin{tabular}{|l|c|c|}
\hline
\textbf{Design Optimized Using} & \textbf{Trajectory Optimization} & \textbf{QP-TSC} \\
\hline
Trajectory Optimization & 1.280 & 1.605 \\
\hline
QP-TSC           & 1.461 & 1.550 \\
\hline
\end{tabular}
\end{table}

The above comparison Table \ref{tab:design_controller_comparison} and the DE results Table \ref{tab:de_runs} show that both the controllers converge to similar link length ratios (1.69 for traj-opt and 1.61 for QP-TSC) but differ in the exact values and their COT is different. The traj-opt performs better than the QP-TSC when run with both optimal designs. Fig. \ref{fig:cot_diff} shows that the traj-opt performs better across most morphologies compared to the QP-TSC.

\subsection{Physical Realizability}
\label{sec:res1d:mass}

The QP-TSC's optimum has a link length of 0.54 and 0.34. The total mass of the legs are 2.2 kg against a total robot mass of 5.2 kg which is implausible once actuator mass is also accounted for. When applying this to a physical pipeline, appropriate leg-mass fraction constraints must be added to avoid this situation.

\section{Planar Jump and Landing Control}
\label{sec:planar}

\subsection{From 1D to 2D}
\label{sec:planar:from1d}
At lift-off, the hip and knee joints possess significant angular velocities. During the flight phase these joint velocities are damped to prepare for landing. Since no external moment acts on the robot after lift-off, reducing the joint angular momentum produces an equal and opposite change in the base angular momentum. Consequently, damping the leg motion causes the base to rotate, whereas constraining the base pitch causes the legs to swing away from the desired landing configuration.
To satisfy both requirements simultaneously, the trajectory optimization generates a
motion in which the centroidal angular momentum is regulated during stance and the leg velocities are redistributed during flight such that the base naturally returns to an upright orientation at touchdown.

\subsection{Trajectory Optimization with Angular Momentum Regulation}
\label{sec:planar:amreg}
The total angular momentum about the center of mass is constrained to zero at lift-off,

\begin{equation}
H(T_{\mathrm{LO}})=0,
\end{equation}

where

\begin{equation}
H=
\sum_i
\left(
I_i\omega_i
+
m_i
(\mathbf{r}_i-\mathbf{r}_{\mathrm{COM}})
\times
\mathbf{v}_i
\right).
\end{equation}

The Horizontal CoM velocity is also constrained to zero for in-place hopping. This can later be modified to achieve forward or backward hopping.

\begin{equation}
\dot{x}_{\mathrm{COM}}(T_{\mathrm{LO}})=0.
\end{equation}

The remaining constraints are similar to the 1D case \ref{sec:framework:trajopt} .

\subsection{Landing: Impact Absorption and LQR Stabilization}
\label{sec:planar:landing}

\begin{figure}
    \centering
    \includegraphics[width=1\linewidth]{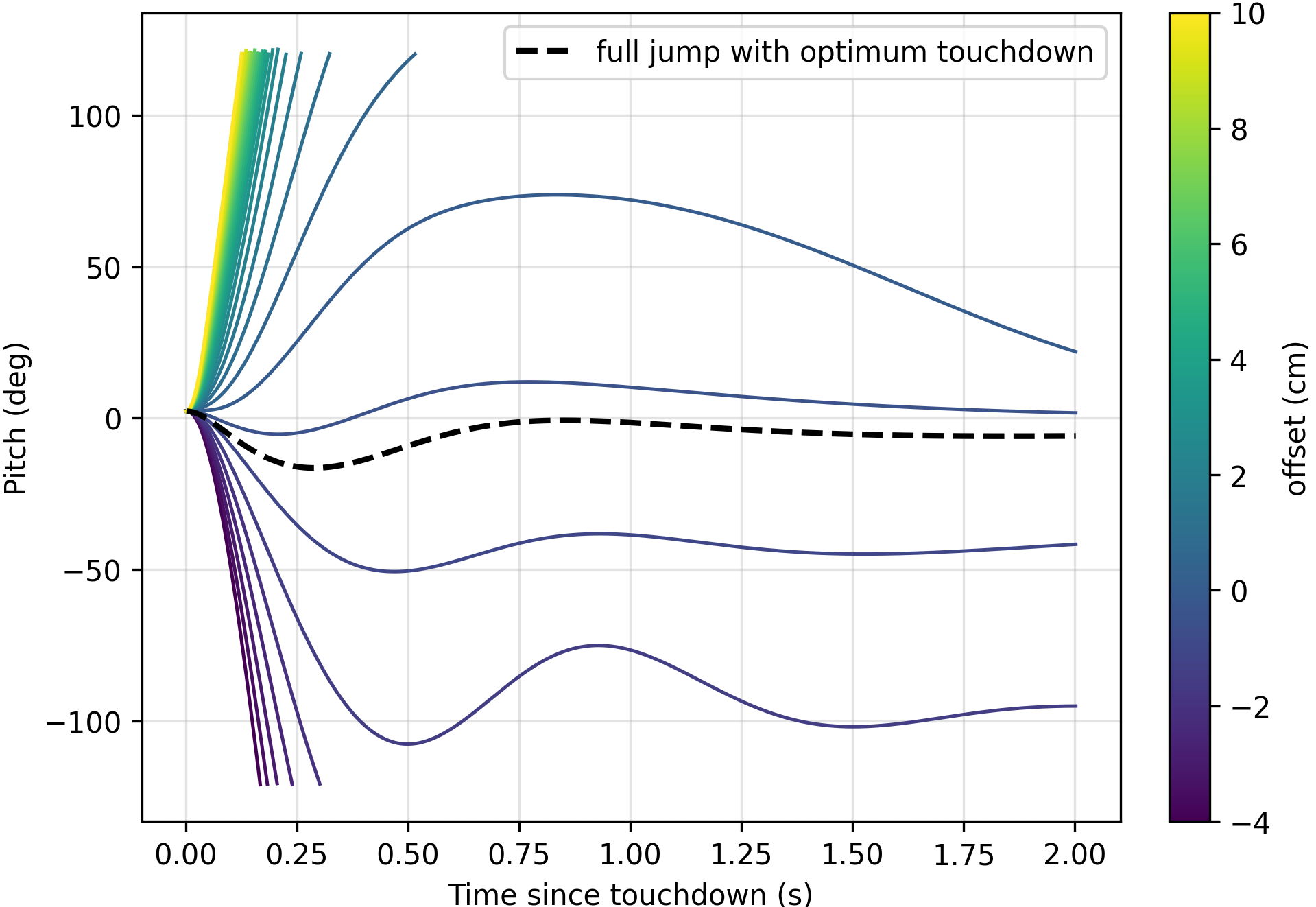}
    \caption{Pitch trajectory for various COM-foot position offset}
    \label{fig:offset}
\end{figure}

The landing phase for the 1D case is very simple : absorb the impact and stabilize which was achieved using a simple joint space PD control. In the 2D case the body pitch also has to be stabilized. The nonlinear dynamics were derived for the reduced model under the assumption that the foot position is fixed and the dynamics were linearized around a nominal landing configuration. An LQR controller was built to absorb the landing impact and stabilize. The LQR works well for stabilization but needs care for absorbing the impact. During the impact the hip torques make the body pitch, to counteract this, the landing foot position needs to be chosen in such a way that it ensures the torques due to ground reaction forces counteract the hip torques that act on the body pitch. The value was chosen after a sweep over the touchdown poses (Fig.\ref{fig:offset}) and the pose with minimum peak body pitch was chosen. The touchdown configuration in the trajectory optimization was chosen to be an extended leg configuration (hip angle = \(35^\circ \)) to ensure there is enough leg stroke to absorb the impact.

\subsection{Region-of-Attraction Analysis}
\label{sec:planar:roa}
A nonlinear simulation sweep was done with varying the touchdown poses to obtain an empirical region of attraction for the LQR landing controller. The actual region is a volume in the 6-dimension space, here two parameters were varied at a time by keeping the others constant. The criteria for convergence was - within \(2^\circ \) of the equilibrium pose and within 0.3 rad/s of zero rate and peak body pitch below \(90^\circ \)

\begin{figure*}[t]
    \centering
    \begin{subfigure}[t]{0.45\textwidth}
        \centering
        \includegraphics[width=\linewidth]{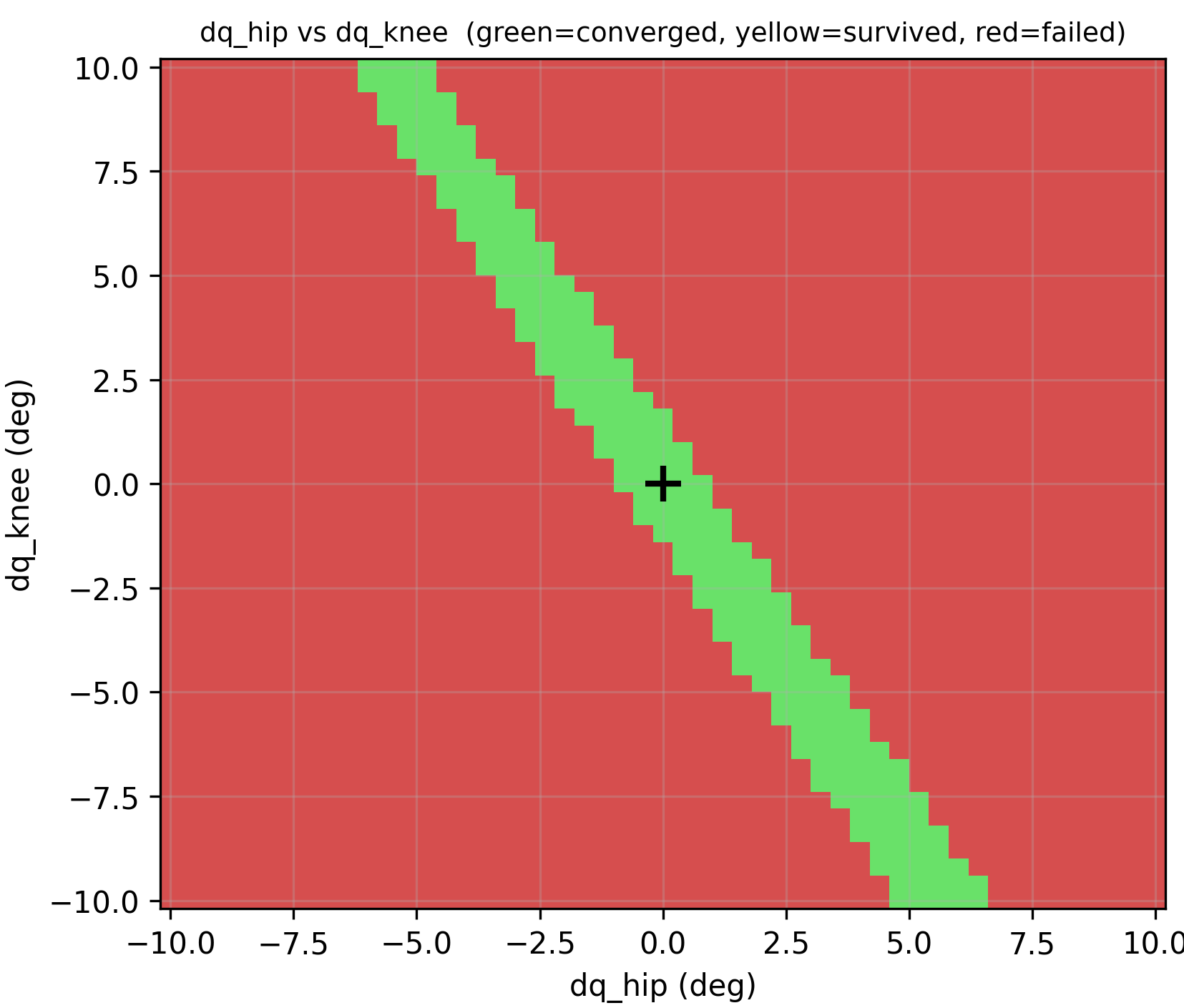}
        \caption{Hip and knee positions perturbed}
        \label{fig:roa1}
    \end{subfigure}\hfill
    \begin{subfigure}[t]{0.45\textwidth}
        \centering
        \includegraphics[width=\linewidth]{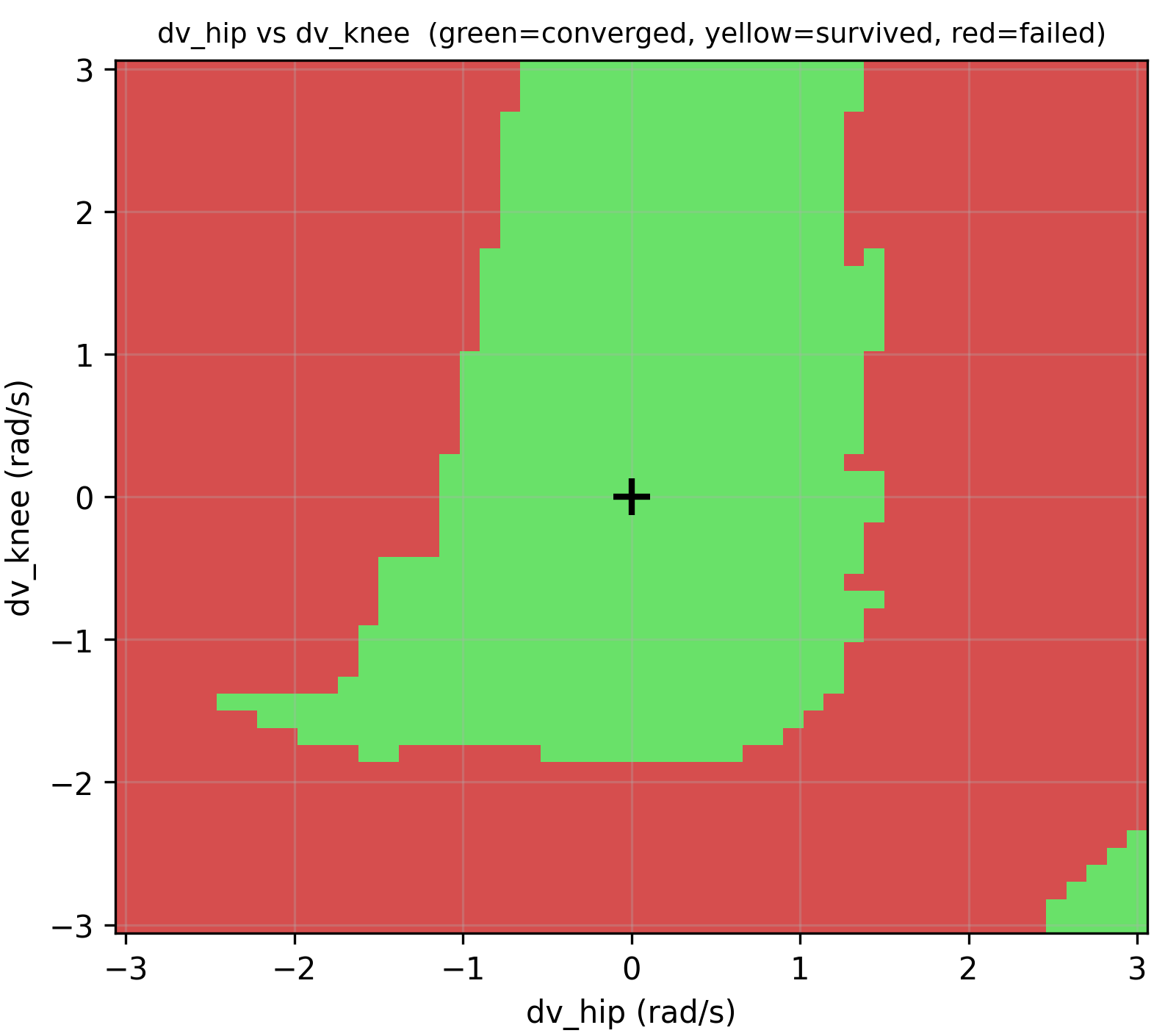}
        \caption{Hip and knee rates perturbed}
        \label{fig:roa2}
    \end{subfigure}
    \caption{Empirical region of attraction of the LQR landing controller.}
    \label{fig:roa}
\end{figure*}

\section{Results: 2D Jump and Landing}
\label{sec:res2d}

\begin{figure}[!h]
    \centering
    \includegraphics[width=1\linewidth]{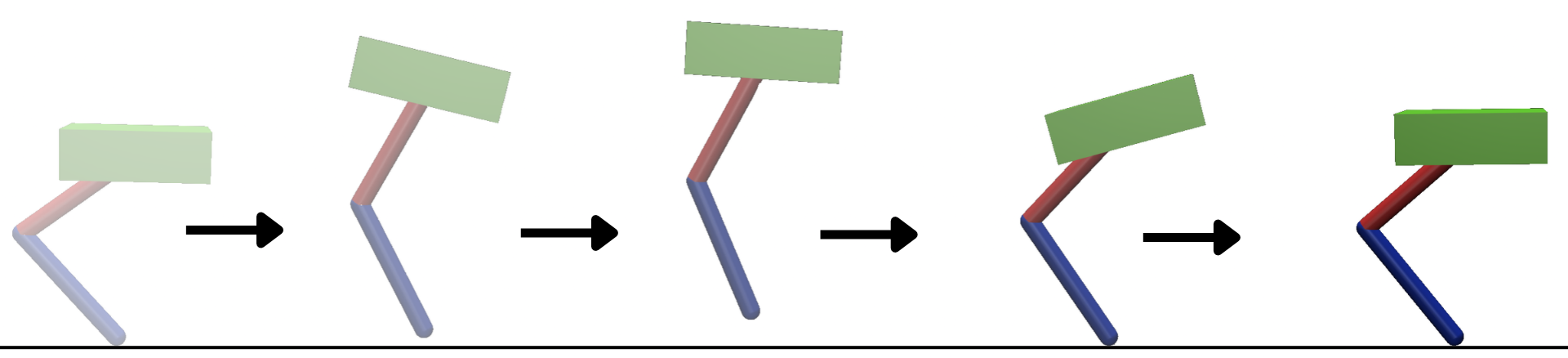}
    \caption{2D in-place jump}
    \label{fig:2djump}
\end{figure}

\subsection{Jump Trajectories}
\label{sec:res2d:am}
\begin{table}[h]
\centering
\caption{2D jump metrics.}
\label{tab:2d_metrics}
\begin{tabular}{|c|c|c|}
\hline
 \textbf{Quantity} & \textbf{In-place} & \textbf{Forward} \\
\hline
Apex CoM height (m) & 0.557 & 0.559 \\
\hline
Liftoff CoM height (m) & 0.353 & 0.333 \\
\hline
Flight time (s) & 0.70 & 0.76 \\
\hline
Horizontal displacement (m) & $\approx 0$ & 0.20 \\
\hline
Peak $|$pitch$|$ before touchdown ($^\circ$) & 43 & 51 \\
\hline
Peak $|$pitch$|$ after touchdown ($^\circ$) & 17 & 79 \\
\hline
\end{tabular}
\end{table}

\begin{figure*}[t]
    \centering
    \begin{subfigure}[t]{0.48\textwidth}
        \centering
        \includegraphics[width=\linewidth]{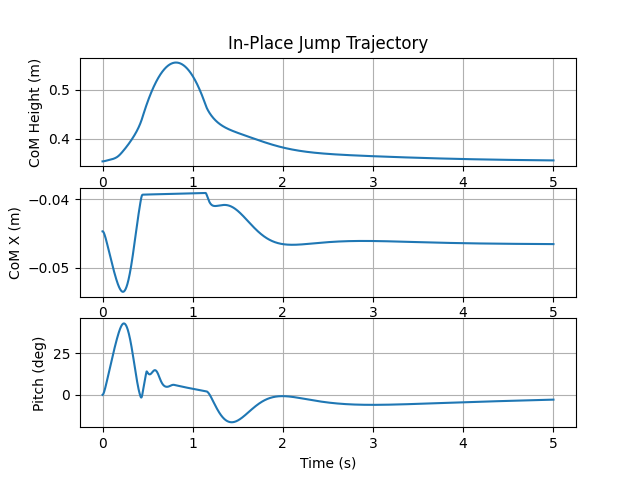}
        \caption{In-place jump}
        \label{fig:intraj}
    \end{subfigure}\hfill
    \begin{subfigure}[t]{0.48\textwidth}
        \centering
        \includegraphics[width=\linewidth]{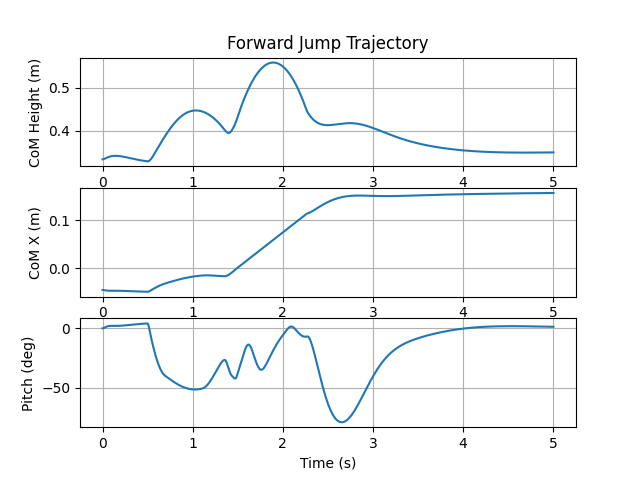}
        \caption{Forward jump}
        \label{fig:fortraj}
    \end{subfigure}
    \caption{2D jump trajectories.}
    \label{fig:2dtraj}
\end{figure*}

\section{Discussion}
\label{sec:disc}

\subsection{1D Co-Design}
The gap between the two controllers is due to the way each controller is built. Trajectory optimization plans the entire stance phase while minimizing energy, while the QP-TSC tracks a liftoff velocity reference and never minimizes energy.

The QP-TSC landscape is also shallower than the trajectory optimization, which is also evident from the fact that the trajectory optimized design loses only 3.5\% under the QP-TSC, while the QP-TSC optimized design loses 14\% under the trajectory optimizer. Both the landscapes have a bottom right triangular white region (Fig \ref{fig:cot_diff}) which corresponds to the violation of static stability pose.

The QP-TSC weights and gains were not re-tuned for each morphology so part of the QP-TSC's performance could be affected due to that.

These findings apply to one task (0.2m jump), one gravity, ideal actuators and fixed initial hip angle and may shift when done for 2D jump.

\subsection{2D jump and Landing Controller}
The planar results show two places where the pitch must be handled: through angular momentum during liftoff and through foot placement at touchdown.

With no external torques during the flight phase, damping the velocities of the leg does make the base rotate and thus zeroing the angular momentum at liftoff was necessary. The base orientation was constrained to be upright with zero velocity at touchdown.

During landing the COM to foot position offset dominates the outcome. Thus the LQR stabilizes a given good touchdown pose but fails to recover from a poor touchdown pose.

The two ROA  Fig.\ref{fig:roa1} and Fig.\ref{fig:roa2} are very different from each other. Position perturbations produce a narrow diagonal band whereas the rate perturbations produce a broad region. This is not enough to conclude that the landing controller is more sensitive to position error than rate errors as they are two different units and cannot be compared directly.

\subsection{Limitations}
\label{sec:disc:lim}

All results are from simulation and no hardware testings were done. The design variables were limited to link lengths, with ideal actuators and the 2D jump was demonstrated on the baseline morphology.

\subsection{Future Work}
\label{sec:disc:future}

Future work will therefore first verify if the co-designed morphology holds for the 2D case, and will also include actuator models and torque-speed curves to make the  problem more realistic. Landing controllers like TV-LQR, MPC or RL-based approaches could be explored for more robust landing. Finally extending the trajectory optimization and task space controller for lunar terrain would test the framework beyond flat terrain.

\section{Conclusion}
\label{sec:concl}

Morphology convergence is controller dependent. Both QP-TSC and trajectory-optimization approaches converged to similar link length ratios (1.61 and 1.69, respectively), but selected different absolute link lengths. This indicates that the choice of controller strongly influences the mechanical parameters that govern COT.

The results also show an asymmetry - the trajectory-optimized morphology achieved a COT of 1.280 with its co-designed controller and 1.605 with the QP-TSC, while the QP-TSC optimized morphology achieved 1.461 with the trajectory optimizer and 1.550 with the QP-TSC. Thus, the trajectory optimizer is the stronger controller in absolute terms, and its advantage persists across morphologies.

The model was further extended to 2D in-place and forward jumping. Angular momentum regulation was incorporated during stance, with LQR control used during the landing phase. Both jumping modes were successfully demonstrated in simulation.

Overall, morphology and controller jointly determine performance, but controller quality is not equivalent across methods, a morphology optimized with a weaker controller can outperform its nominal design when paired with a stronger controller.

\appendices

\section{Implementation Details}
\label{app:imp}
\subsection{QP-TSC parameters}
\label{app:imp:qp}
\begin{itemize}[leftmargin=*,itemsep=0pt,topsep=2pt,parsep=0pt]
    \item For the stance phase - Kp = [0,0,0] and Kd = [300,0,0]
    \item For the flight phase - Kp = [0,50,50] and Kd = [0,10,10]
    \item For Stance phase - W = diag(300, 0, 0)
    \item For Flight phase - W = diag(0, 100, 100)
    \item Phase switching is contact based with debounce
\end{itemize}

\subsection{Trajectory optimization}
\label{app:imp:traj}
\begin{itemize}[leftmargin=*,itemsep=0pt,topsep=2pt,parsep=0pt]
    \item Stance duration is a free variable
    \item 1D case is discretized into 75 nodes
    \item 2D case stance is discretized into 75 nodes and flight phase is discretized into 60 nodes
\end{itemize}

\subsection{Differential Evolution Parameters}
\label{app:imp:de}
\begin{itemize}[leftmargin=*,itemsep=0pt,topsep=2pt,parsep=0pt]
    \item Search Bounds - 0.15 m to 0.6 m
    \item Variant - 2 (random selection + exponential crossover)
    \item Mutation factor(F) = 0.8 and cross-over rate(cr) = 0.9
    \item Population size = 50
    \item No. of generations = 75
    \item No. of different seeds used = 5
\end{itemize}

\section{LQR Derivation and Gains}
\label{app:lqr}

The foot-pinned monoped was linearized about its static equilibrium
configuration for LQR controller design. The generalized coordinates are

\begin{equation}
\mathbf{q} =
\begin{bmatrix}
\theta_0 & \theta_1 & \theta_2
\end{bmatrix}^{T},
\end{equation}

where $\theta_0$ is the body pitch angle, $\theta_1$ is the hip angle,
and $\theta_2$ is the knee angle. The hip and knee are actuated, while
the body pitch is unactuated. Thus,

\begin{equation}
\mathbf{u} =
\begin{bmatrix}
\tau_1 & \tau_2
\end{bmatrix}^{T},
\qquad
B_u =
\begin{bmatrix}
0 & 0\\
1 & 0\\
0 & 1
\end{bmatrix}.
\end{equation}

The system dynamics were obtained using the Lagrangian formulation and the resulting equations of motion are written as

\begin{equation}
M(\mathbf q)\ddot{\mathbf q}
+C(\mathbf q,\dot{\mathbf q})\dot{\mathbf q}
+G(\mathbf q)
=
B_u\mathbf u.
\end{equation}

\subsection{Equilibrium Configuration}

The linearization is performed about a static equilibrium, for which

\begin{equation}
\dot{\mathbf q}_0=\mathbf 0,
\qquad
\ddot{\mathbf q}_0=\mathbf 0.
\end{equation}

The touchdown configuration is

\begin{align}
\theta_{1,0} &= 55^\circ,\\
\theta_{2,0} &= -97.0837^\circ.
\end{align}

The body pitch required to satisfy the unactuated equilibrium condition
$G_1(\mathbf q_0)=0$ is

\begin{equation}
\theta_{0,0}=0^\circ.
\end{equation}

Therefore,

\begin{equation}
\mathbf q_0=
\begin{bmatrix}
0 & 55^\circ & -97.0837^\circ
\end{bmatrix}^{T},
\end{equation}

with the corresponding equilibrium torques

\begin{equation}
\mathbf u_0=
\begin{bmatrix}
0\\
1.3478
\end{bmatrix}
\mathrm{N\,m}.
\end{equation}

\subsection{Linearized Model}

Defining the state as

\begin{equation}
\mathbf x =
\begin{bmatrix}
\mathbf q\\
\dot{\mathbf q}
\end{bmatrix},
\end{equation}

the linearized dynamics are

\begin{equation}
\delta\dot{\mathbf x}
=
A\delta\mathbf x+B\delta\mathbf u,
\end{equation}

The resulting numerical state-space matrices are

\begin{equation}
A=
\begin{bmatrix}
0 & 0 & 0 & 1 & 0 & 0\\
0 & 0 & 0 & 0 & 1 & 0\\
0 & 0 & 0 & 0 & 0 & 1\\
0 & 0 & 0 & 0 & 0 & 0\\
3.8052 & 3.8052 & 0.5362 & 0 & 0 & 0\\
0.7222 & 0.7222 & 3.6521 & 0 & 0 & 0
\end{bmatrix},
\end{equation}

and

\begin{equation}
B=
\begin{bmatrix}
0 & 0\\
0 & 0\\
0 & 0\\
-40.0000 & 0\\
43.4639 & -3.1046\\
-3.1046 & 5.5517
\end{bmatrix}.
\end{equation}

The controllability matrix has full rank,

\begin{equation}
\operatorname{rank}
\begin{bmatrix}
B & AB & A^2B & \cdots & A^5B
\end{bmatrix}
=6,
\end{equation}

indicating that the linearized system is fully controllable.

\subsection{LQR Controller}

The LQR controller was designed using

\begin{equation}
Q=
\operatorname{diag}(500,10,10,500,5,5),
\qquad
R=
\operatorname{diag}(1500,1500).
\end{equation}

The resulting feedback gain is

\begin{equation}
K_{\mathrm{LQR}}=
\begin{bmatrix}
48.83 & 48.80 & 29.31 & 24.13 & 23.78 & 13.65\\
49.60 & 49.02 & 30.64 & 24.73 & 23.83 & 14.34
\end{bmatrix}
\end{equation}

The control law is

\begin{equation}
\mathbf u
=
\mathbf u_0
-
K_{\mathrm{LQR}}
(\mathbf x-\mathbf x_0).
\end{equation}

The closed-loop eigenvalues are $\{-1.0055, -1.6688, -1.8803, -2.0672, -2.1098, -23.2090\}$.

\section{Additional Results}
\label{app:extra}
\subsection{DE results for both conservative and extended bounds}

In preliminary runs with the bounds set to 0.15 m to 0.3 m, both controllers converged to nearly the same design, with $l_1$ hitting the upper bound of 0.300 m. The mean optima over five seeds were $l_2 = 0.221$ m (COT 1.61) for the QP-TSC and $l_2 = 0.225$ m (COT 1.35) for trajectory optimization. Since the first link was hitting the bound, the search range was extended to 0.15 m to 0.6 m to find the true minima.

\begin{table}[h]
\centering
\caption{Five independent DE runs (extended bounds).}
\label{tab:de_runs_both}
\resizebox{\columnwidth}{!}{%
\begin{tabular}{|c|ccc|ccc|}
\hline
 & \multicolumn{3}{c|}{\textbf{Traj-opt}} & \multicolumn{3}{c|}{\textbf{QP-TSC}} \\
\textbf{Run} & $l_1$ & $l_2$ & COT & $l_1$ & $l_2$ & COT \\
\hline
1 & 0.3562 & 0.2114 & 1.2824 & 0.5452 & 0.3387 & 1.5499 \\
2 & 0.3564 & 0.2116 & 1.2825 & 0.5470 & 0.3401 & 1.5498 \\
3 & 0.3571 & 0.2119 & 1.2824 & 0.5515 & 0.3438 & 1.5498 \\
4 & 0.3568 & 0.2119 & 1.2830 & 0.5448 & 0.3383 & 1.5498 \\
5 & 0.3575 & 0.2122 & 1.2825 & 0.5295 & 0.3319 & 1.5497 \\
\hline
\textbf{Mean} & \textbf{0.36} & \textbf{0.21} & \textbf{1.28} & \textbf{0.54} & \textbf{0.34} & \textbf{1.55} \\
\hline
\end{tabular}}
\end{table}

\begin{figure}[t]
    \centering
    \begin{subfigure}[t]{0.9\linewidth}
        \centering
        \includegraphics[width=\linewidth]{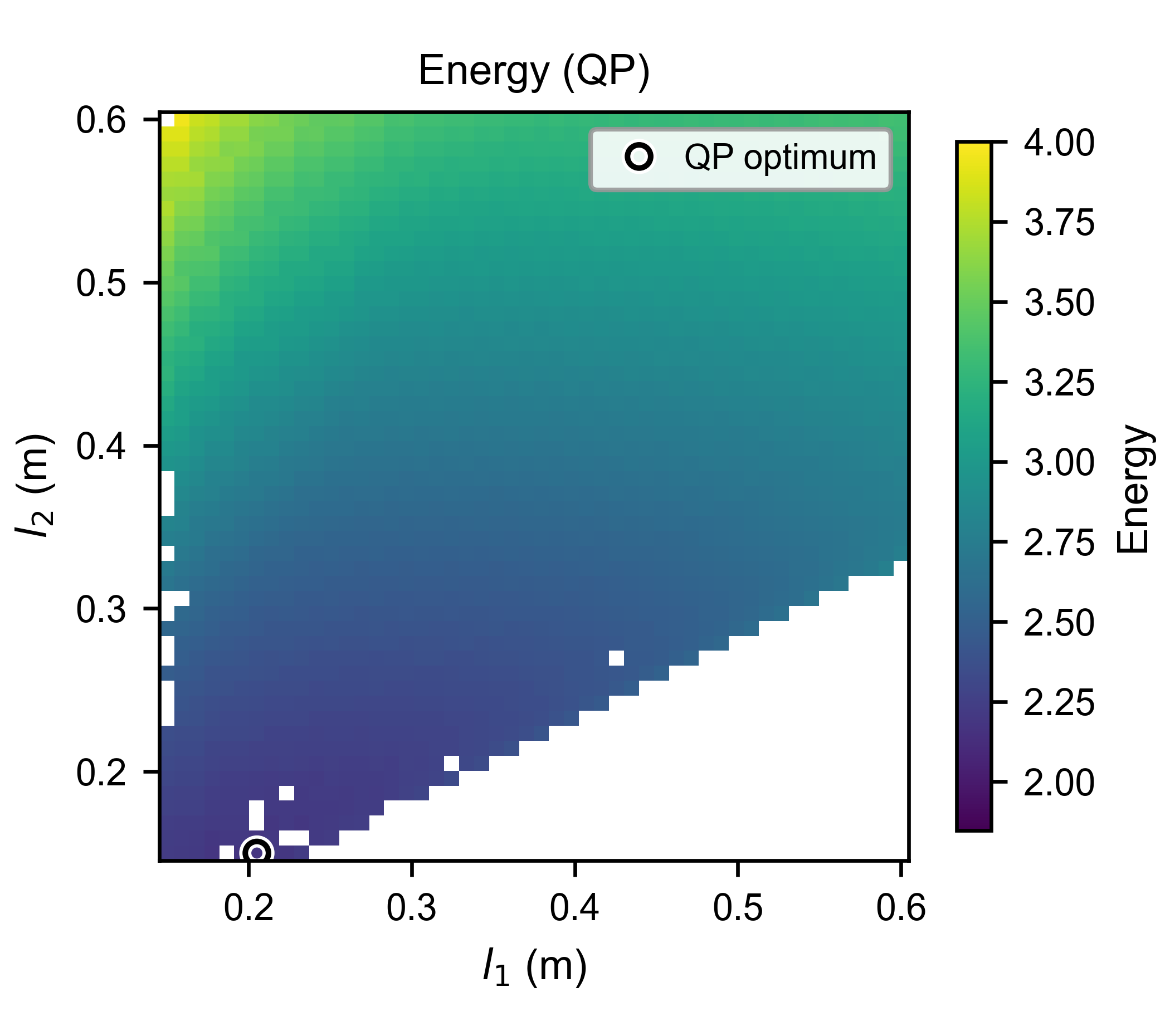}
        \caption{QP-TSC}
        \label{fig:enerqp}
    \end{subfigure}

    \vspace{4pt}
    \begin{subfigure}[t]{0.9\linewidth}
        \centering
        \includegraphics[width=\linewidth]{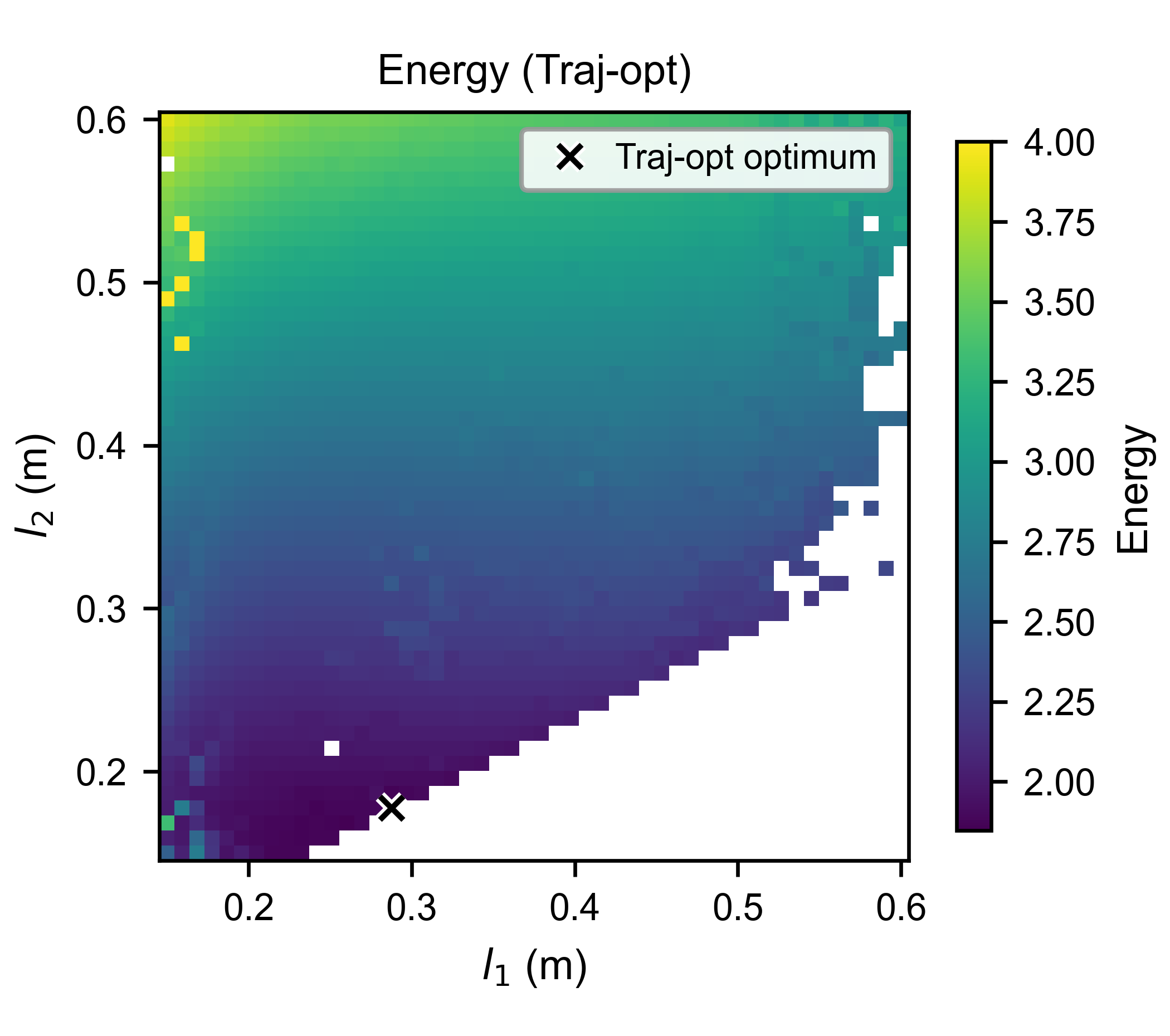}
        \caption{Trajectory optimization}
        \label{fig:enertraj}
    \end{subfigure}
    \caption{Energy landscapes over link lengths.}
    \label{fig:energy}
\end{figure}

The Fig.\ref{fig:enerqp} and Fig.\ref{fig:enertraj} support the claim that using raw energy will bias the optimizer toward smaller link lengths.

\bibliographystyle{IEEEtran}
\bibliography{refs}   

@inproceedings{8794136,
  author={Arm, Philip and Zenkl, Radek and Barton, Patrick and Beglinger, Lars and Dietsche, Alex and Ferrazzini, Luca and Hampp, Elias and Hinder, Jan and Huber, Camille and Schaufelberger, David and Schmitt, Felix and Sun, Benjamin and Stolz, Boris and Kolvenbach, Hendrik and Hutter, Marco},
  booktitle={2019 International Conference on Robotics and Automation (ICRA)}, 
  title={SpaceBok: A Dynamic Legged Robot for Space Exploration}, 
  year={2019},
  volume={},
  number={},
  pages={6288-6294},
  doi={10.1109/ICRA.2019.8794136}}

@INPROCEEDINGS{8967552,
  author={Kolvenbach, Hendrik and Hampp, Elias and Barton, Patrick and Zenkl, Radek and Hutter, Marco},
  booktitle={2019 IEEE/RSJ International Conference on Intelligent Robots and Systems (IROS)}, 
  title={Towards Jumping Locomotion for Quadruped Robots on the Moon}, 
  year={2019},
  volume={},
  number={},
  pages={5459-5466},
  doi={10.1109/IROS40897.2019.8967552}}

@misc{spiridonov2024spacehoppersmallscaleleggedrobot,
      title={SpaceHopper: A Small-Scale Legged Robot for Exploring Low-Gravity Celestial Bodies}, 
      author={Alexander Spiridonov and Fabio Buehler and Moriz Berclaz and Valerio Schelbert and Jorit Geurts and Elena Krasnova and Emma Steinke and Jonas Toma and Joschua Wuethrich and Recep Polat and Wim Zimmermann and Philip Arm and Nikita Rudin and Hendrik Kolvenbach and Marco Hutter},
      year={2024},
      eprint={2403.02831},
      archivePrefix={arXiv},
      primaryClass={cs.RO},
      url={https://arxiv.org/abs/2403.02831}, 
}

@misc{olsen2025olympusjumpingquadrupedplanetary,
      title={Olympus: A Jumping Quadruped for Planetary Exploration Utilizing Reinforcement Learning for In-Flight Attitude Control}, 
      author={Jørgen Anker Olsen and Grzegorz Malczyk and Kostas Alexis},
      year={2025},
      eprint={2503.03574},
      archivePrefix={arXiv},
      primaryClass={cs.RO},
      url={https://arxiv.org/abs/2503.03574}, 
}

@misc{hari2026dynamicmodelingattitudecontrol,
      title={Dynamic Modeling and Attitude Control of a Reaction-Wheel-Based Low-Gravity Bipedal Hopper}, 
      author={Shriram Hari and M Venkata Sai Nikhil and R Prasanth Kumar},
      year={2026},
      eprint={2603.10670},
      archivePrefix={arXiv},
      primaryClass={cs.RO},
      url={https://arxiv.org/abs/2603.10670}, 
}

@misc{olsen2023martianlavatubeexploration,
      title={Martian Lava Tube Exploration Using Jumping Legged Robots: A Concept Study}, 
      author={Jørgen Anker Olsen and Kostas Alexis},
      year={2023},
      eprint={2310.14876},
      archivePrefix={arXiv},
      primaryClass={cs.RO},
      url={https://arxiv.org/abs/2310.14876}, 
}

@article{Rudin_2022,
   title={Cat-Like Jumping and Landing of Legged Robots in Low Gravity Using Deep Reinforcement Learning},
   volume={38},
   ISSN={1941-0468},
   url={http://dx.doi.org/10.1109/TRO.2021.3084374},
   DOI={10.1109/tro.2021.3084374},
   number={1},
   journal={IEEE Transactions on Robotics},
   publisher={Institute of Electrical and Electronics Engineers (IEEE)},
   author={Rudin, Nikita and Kolvenbach, Hendrik and Tsounis, Vassilios and Hutter, Marco},
   year={2022},
   month=Feb, pages={317–328} }

@INPROCEEDINGS{Ha-RSS-17, 
    AUTHOR    = {Sehoon Ha AND Stelian Coros AND Alexander Alspach AND Joohyung Kim AND Katsu Yamane}, 
    TITLE     = {Joint Optimization of Robot Design and Motion Parameters using the Implicit Function Theorem}, 
    BOOKTITLE = {Proceedings of Robotics: Science and Systems}, 
    YEAR      = {2017}, 
    ADDRESS   = {Cambridge, Massachusetts}, 
    MONTH     = {July}, 
    DOI       = {10.15607/RSS.2017.XIII.003} 
}

@misc{dinev2022versatilecodesignapproachdynamic,
      title={A Versatile Co-Design Approach For Dynamic Legged Robots}, 
      author={Traiko Dinev and Carlos Mastalli and Vladimir Ivan and Steve Tonneau and Sethu Vijayakumar},
      year={2022},
      eprint={2103.04660},
      archivePrefix={arXiv},
      primaryClass={cs.RO},
      url={https://arxiv.org/abs/2103.04660}, 
}

@misc{schaff2018jointlylearningconstructcontrol,
      title={Jointly Learning to Construct and Control Agents using Deep Reinforcement Learning}, 
      author={Charles Schaff and David Yunis and Ayan Chakrabarti and Matthew R. Walter},
      year={2018},
      eprint={1801.01432},
      archivePrefix={arXiv},
      primaryClass={cs.RO},
      url={https://arxiv.org/abs/1801.01432}, 
}

@article{10.1115/1.4052463,
    author = {Bravo-Palacios, Gabriel and Grandesso, Gianluigi and Prete, Andrea Del and Wensing, Patrick M.},
    title = {Robust Co-Design: Coupling Morphology and Feedback Design Through Stochastic Programming },
    journal = {Journal of Dynamic Systems, Measurement, and Control},
    volume = {144},
    number = {2},
    pages = {021007},
    year = {2021},
    month = {11},
    issn = {0022-0434},
    doi = {10.1115/1.4052463},
    url = {https://doi.org/10.1115/1.4052463},
    eprint = {https://asmedigitalcollection.asme.org/dynamicsystems/article-pdf/144/2/021007/6799030/ds_144_02_021007.pdf},
}

@INPROCEEDINGS{9560988,
  author={Fadini, G. and Flayols, T. and Del Prete, A. and Mansard, N. and Souères, P.},
  booktitle={2021 IEEE International Conference on Robotics and Automation (ICRA)}, 
  title={Computational design of energy-efficient legged robots: Optimizing for size and actuators}, 
  year={2021},
  volume={},
  number={},
  pages={9898-9904},
  doi={10.1109/ICRA48506.2021.9560988}}

@INPROCEEDINGS{11203144,
  author={Singh, Aman and Mishra, Aastha and Kapa, Deepak and Joshi, Suryank and Kolathaya, Shishir},
  booktitle={2025 IEEE-RAS 24th International Conference on Humanoid Robots (Humanoids)}, 
  title={A Co-Design Framework for Energy-Aware Monoped Jumping with Detailed Actuator Modeling}, 
  year={2025},
  volume={},
  number={},
  pages={1012-1018},
  doi={10.1109/Humanoids65713.2025.11203144}}

@misc{choi2022computationregionsattractionhybrid,
      title={Computation of Regions of Attraction for Hybrid Limit Cycles Using Reachability: An Application to Walking Robots}, 
      author={Jason J. Choi and Ayush Agrawal and Koushil Sreenath and Claire J. Tomlin and Somil Bansal},
      year={2022},
      eprint={2201.08538},
      archivePrefix={arXiv},
      primaryClass={cs.RO},
      url={https://arxiv.org/abs/2201.08538}, 
}

@article{osti_10182618,
title = {Control policies for a large region of attraction for dynamically balancing legged robots: a sampling-based approach}, url = {https://par.nsf.gov/biblio/10182618}, DOI = {10.1017/S0263574720000211}, abstractNote = {The popular approach of assuming a control policy and then finding the largest region of attraction (ROA) (e.g., sum-of-squares optimization) may lead to conservative estimates of the ROA, especially for highly nonlinear systems. We present a sampling-based approach that starts by assuming a ROA and then fi nds the necessary control policy by performing trajectory optimization on sampled initial conditions. Our method works with black-box models, produces a relatively large ROA, and ensures exponential convergence of the initial conditions to the periodic motion. We demonstrate the approach on a model of hopping and include extensive verification and robustness checks.}, journal = {Robotica}, author = {Bhounsule, P. A. and Kim, M. and Alaeddini, A.}, }

@INPROCEEDINGS{Posa-RSS-17, 
    AUTHOR    = {Michael Posa AND Twan Koolen AND Russ Tedrake}, 
    TITLE     = {Balancing and Step Recovery Capturability via Sums-of-Squares Optimization}, 
    BOOKTITLE = {Proceedings of Robotics: Science and Systems}, 
    YEAR      = {2017}, 
    ADDRESS   = {Cambridge, Massachusetts}, 
    MONTH     = {July}, 
    DOI       = {10.15607/RSS.2017.XIII.032} 
}

\end{document}